\documentclass{article}

\usepackage{microtype}
\usepackage{graphicx}
\usepackage{subfigure}
\usepackage{booktabs} 
\usepackage{braket}
\usepackage{amsmath}

\usepackage{hyperref}

\usepackage[accepted]{icml2019}

\icmltitlerunning{Perturbation theory on Variational Autoencoders}

\begin{document}

\twocolumn[
\icmltitle{Perturbation theory approach to study the \\
           latent space degeneracy of Variational Autoencoders}


\icmlsetsymbol{equal}{*}

\begin{icmlauthorlist}
\icmlauthor{Helena Andr\'es-Terr\'e}{equal,cam}
\icmlauthor{Pietro Li\'o}{cam}

\end{icmlauthorlist}

\icmlaffiliation{cam}{Department of Computer Science, University of Cambridge, Cambridge, United Kingdom}

\icmlcorrespondingauthor{Helena Andr\'es-Terr\'e}{ha376@cam.ac.uk}

\icmlkeywords{Machine Learning, Physics}

\vskip 0.3in
]



\printAffiliationsAndNotice{} 

%
%
%

\begin{abstract}

The use of Variational Autoencoders in different Machine Learning tasks has drastically increased in the last years. They have been developed as denoising, clustering and generative tools, highlighting a large potential in a wide range of fields. Their embeddings are able to extract relevant information from highly dimensional inputs, but the converged models can differ significantly and lead to degeneracy on the latent space. We leverage the relation between theoretical physics and machine learning to explain this behaviour, and introduce a new approach to correct for degeneration by using perturbation theory. The re-formulation of the embedding as multi-dimensional generative distribution, allows mapping to a new set of functions and their corresponding energy spectrum. We optimise for a perturbed Hamiltonian, with an additional energy potential that is related to the unobserved topology of the data. Our results show the potential of a new theoretical approach that can be used to interpret the latent space and generative nature of unsupervised learning, while the energy landscapes defined by the perturbations can be further used for modelling and dynamical purposes.

\end{abstract}
\section{Introduction}
\label{introduction}


The development of unsupervised learning techniques in the field of Artificial Intelligence (AI) has highlighted the need of a theoretical framework to interpret and analyse the results \cite{engel2001statistical, nishimori2001statistical,mezard2009information}. Unsupervised learning often has an exploratory objective, with some of its goals being to explore new patterns or unknown features among the data. The broad nature of its usage becomes a challenge in the evaluation stage, where there is not a consensus among methods to compare and analyse the results. 
Variational Autoencoders (VAEs) \cite{Kingma2013auto, rezende2014stochastic} are part of this collective, where they have been given multiple functions throughout their development.  

One of the most common goals of unsupervised learning is to find structures or similarities among the data, often combined with clustering analysis. Even though the labels are unknown, most algorithms require prior assumptions or bounded solutions, restricting the number of configurations allowed to be explored. 

VAEs provide a lower dimensional representation of the data, in the form of a set of generative functions capable to reproduce and reconstruct the original input. But their own composition, being made out of combined deep neural networks, makes them very prone to suffer from model degeneracy \cite{degenerate_Zheng, kaplan2016instability}.

Inspired by perturbation analysis in quantum physics, we have developed an approach to unveil structures and the energy spectrum encoded in the data by using the generative functions extracted from a VAE. The embeddings provide a set of observations of the true particles, which collapse to a spectrum of energy states when exposed to a certain operator. 

Such energies correspond to the different states of the system, providing a new interpretation of the data. 
Our analysis is also able to define one or several potential landscapes associated to the system, which can be used for further exploration of the topology and dynamics behind the data.

\begin{figure*}[t]
\vskip 0.2in
\begin{center}
    \centerline{ \includegraphics[width=1.\textwidth]{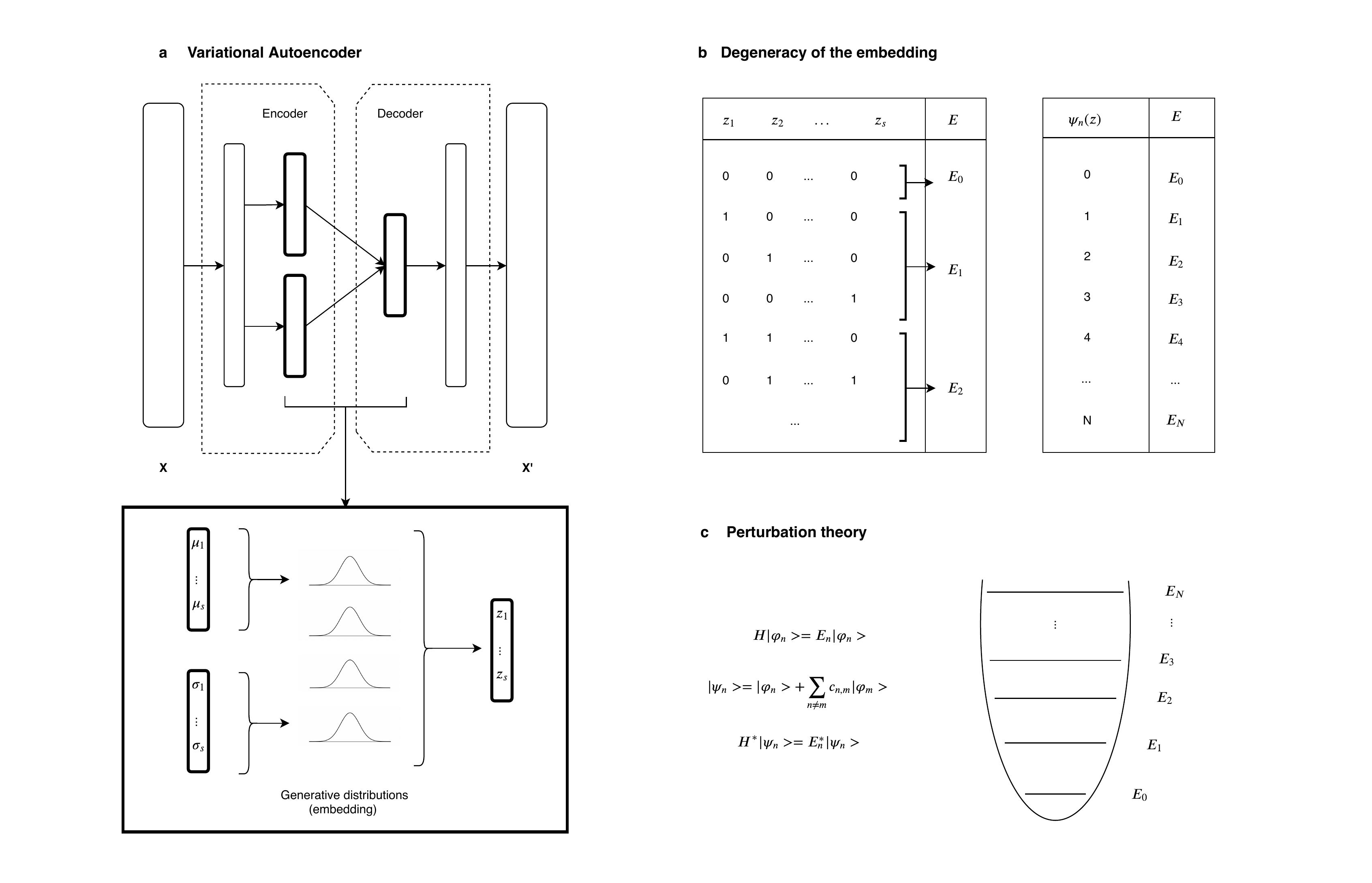}}
\caption{\textbf{Diagram of the perturbation analysis approach to degeneracy on VAEs.} \textbf{(a)} The VAE embedding is a multi-dimensional generative function, where each component $z$ is fitted to a Gaussian distribution. The latent space provides a lower dimensional representation of the data, and often encodes for relevant features and properties among the samples. \textbf{(b)}  The multi-dimensional nature of the embedding leads to degeneracy among models. All solutions reached by convergence of the model under different initialisations, or even different hyper-parameters, should only depend on the input data and therefore be the same among comparable replicas. Instead, for a certain energy or label assigned to the particles, we may obtain different embeddings on parallel models due to symmetries among the latent components. \textbf{(c)} We use a perturbation theory approach to unveil a spectrum of energies that can be associated to the samples, building a more general approach to interpret the latent space generated from VAEs. By applying a certain perturbed Hamiltonian over the generated functions, we find the energy spectrum that represents our system. The perturbed functions can be expressed as a linear combination of the unperturbed ones, where the coefficients $c_{n,m}$ characterise their separability}
\label{fig:general_diagram}
\end{center}
\vskip -0.2in
\end{figure*}


\subsection{Variational Autoencoders}
The Variational Autoencoder, proposed by Kingma and Welling \cite{Kingma2013auto}, uses stochastic inference to approximate the latent variables as probability distributions. The original input can then be reconstructed from the latent space, which also captures relevant features from the data. It is scalable to large datasets, and can deal with intractable posterior distributions by fitting an approximate inference or recognition model, using a reparametrised variational lower bound estimator.
They have been broadly tested and used for data compression or dimensionality reduction, their adaptability and potential to handle non-linear behaviours has made them particularly well fitted to work with complex data.

Variational Autoencoders are built upon a probabilistic framework where the high dimensional data or input $X = x_{i}$ with $i=1...N$ is drawn from a continuous random variable $x$ with distribution $p_{data}(x)$. 
It assumes that the natural data $X$ lies in a lower dimensional space, that can be characterised by an unobserved continuous random variable $z$ where $z \in S$.

In the Bayesian approach, the prior $p_{\theta}(z)$ and conditional or likelihood $p_{\theta}(x|z)$ come from a family of parametric distributions, with Probability Density Functions (PDFs) differentiable almost everywhere with respect to both $\theta$ and $z$. The true parameters $\theta$ and the values of the latent variables $z$ are unknown to us, but the VAE approximates the often intractable true posterior $p_{\theta}(x|z)$ by using a recognition model $q_{\phi}(x|z)$ and the learned parameters $\phi$ represented by the weights of a neural network.

The VAE embedding constitutes a set of multivariate gaussians $\{Z_{i}\}$ or generative functions that represent our data. 

$$Z = \frac{exp(-\frac{1}{2}(z^{*} - z_{\mu})^{T}\Sigma^{-1}(z^{*} - z_{\mu}))}{\sqrt{(2\pi)^{k}|\Sigma |}} $$

Each sample can be seen as an observation of a generative function defined by the embedding $Z_{i}$
, where $z_{\mu}$ and $\Sigma$ are defined by the outputs of the encoder.  The input of the decoder $z^{*}$ is obtained from the reparametrisation trick.

VAEs are optimised to generate a positive definite covariance matrix $\Sigma$ by maximising the disentanglement between components. But despite the convergence of the models, that is not always the case and one can find $\Sigma$ that are not full rank, meaning that the multivariate normal distribution is degenerate and does not have a density function. 

Selecting a subset of the coordinates of $rank(\Sigma)$, or a different base measure where the covariance is positive definite should end the degeneracy. 
If the choice of new coordinates only aims to restrict the functions into a $rank(\Sigma)$-dimensional affine subspace, which can supports the Gaussian distributions, nothing prevents the loss of information. 
For this reason, we have developed a Multilayer Perceptron approach to generate a function $\psi_{n}(Z)$, accounting for the degeneracy of $Z$. It is able to define a unique set of generative functions and encode the relevant features of the data.   

\subsection{Perturbation Theory}
\label{sec:Perturbation Theory}

In order to find the energy states of the system, we use a perturbation theory approach, broadly used in atomic physics, condensed matter and particle physics to solve problems in quantum mechanics. It uses a scheme of successive correction to the zero-field values of energy levels and wavefunctions.

Given a Hamiltonian $H$ with known eigenkets and eigenvalues $H \ket{\psi_{n}} = E_{n} \ket{\psi_{n}} $, one can study how these eigenstates and eigen-energies change when small perturbations are added.

\begin{gather*}
\label{eq:Hamiltonian_pert}
H \ket{\psi_{n}} =  E_{n} \ket{\psi_{n}} \\
(H^{0} + H^{1}) \ket{\psi_{n}} \approx (E^{0}_{n}+E^{1}_{n}) \ket{\psi_{n}}
\end{gather*}

We have used the first order approximation of the expansions $ H = \sum_{i}^{m}{H^{i}} $ and $ E_{n} = \sum_{i}^{m}\lambda^{i}{E_{n}^{i}} $. The VAE generative functions $\ket{\psi_{n}}$ can be expressed as a combination of a ground state and a perturbation term 

$$\ket{\psi_{n}} = \ket{\varphi_{n}} + \sum_{m \neq n} c_{m, n} \ket{\varphi_{m}}$$

Where $ \ket{\varphi_{m}}$ are the wave-functions derived from the unperturbed problem. 

Using an unperturbed Hamiltonian $H^{0}$, equivalent to the kinetic term of the well studied particle in a box example in quantum physics 

$$H^{0} = - A \frac{d^{2}}{dz^{2}}$$ 

The unperturbed wave-functions and energies can be found by solving the Schr\"odinger equation, which gives 

$$E^{0}_{n} = (\frac{\Pi}{L})^{2}  A  n^{2}$$ $$\varphi_{n} = \sqrt{\frac{2}{L}} sin (\frac{\Pi  n}{L}  z)$$

We have added a first order perturbation of the Hamiltonian in the form of 

\begin{equation}
\label{eq:pert_V}
H^{1} = V(z) = sin(2\pi t z)
\end{equation}

With $t$ evenly distributed minima on the $z$ domain. The choice of potential $V(z)$ needs to be made according to the data nature or structure. 

The perturbed energies $E_{n}^{1}$ and wave functions can be derived using

\begin{gather*}
\label{eq:pert_energies_wavefunct}
E_{n}^{1} =  \bra{\varphi_{n}} H^{1} \ket{\varphi_{n}} \\
\ket{\psi_{n}} = \ket{\varphi_{n}} + \sum_{m \neq n} c_{m, n} \ket{\varphi_{m}} \\
c_{m, n} = \bra{\varphi_{m}} H^{1} \ket{\varphi_{n}}
\end{gather*}

Given all the perturbed and unperturbed forms of energies and wave functions, one can extract a set of energies $E_{n}$ and wave-functions that $\ket{\psi_{n}}$ that uniquely describe the system. 

\begin{figure*}[h]
\vskip 0.2in
\begin{center}

    \includegraphics[width=1.\textwidth]{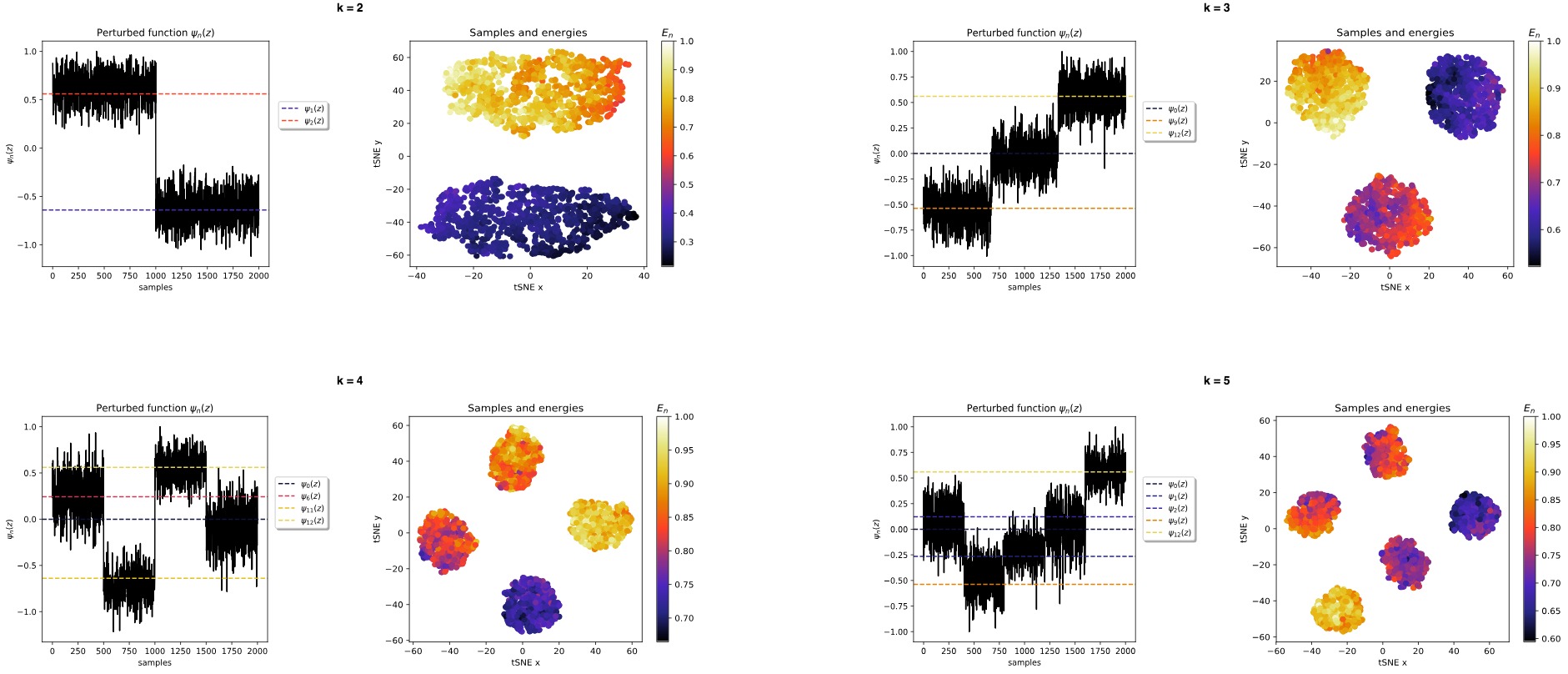}

\caption{\textbf{Functions $\psi_{n}(z)$ and spectrum of energy values for highly dimensional synthetic data, with clusters $\textbf{k=\{2,3,4,5\}}$.} A function $\psi_{n}(z)$ is assigned to each group, allowing to measure the energy difference among them in a fully unsupervised manner. Depicted are for each $k$, the observations of functions $\psi_{n}(z)$ corresponding to the different groups (left). The contiguous plots are tSNE visualisations of the embeddings generated (right), which prove their ability to separate samples according to their labels. The samples are coloured according to the converged energy spectrum.  }
\label{fig:energies_potential_schrodi}
\end{center}
\vskip -0.2in
\end{figure*}




\section{Implementation}
\label{implementation}


We tested our approach on artificially generated data using sklearn toolkit random sample generator \cite{scikit-learn}. The multi-class datasets are created by 	allocating each class one or multiple normally-distributed clusters of points, while adding some noise in the form of correlated, redundant and uninformative features. Pytorch \cite{paszke2017automatic} version 0.4.1. was used to build and train the VAEs and Neural Networks.
The $|X| = D$ dimensional datasets are then used to train a VAE, that will generate an embedding $Z$ which eliminates the noise and preserves the relevant information from the original input. 

The degenerate nature of the embeddings means that different models or embeddings obtained from a VAE using the same dataset may lead to different results. That creates a problem in terms of interpretation and generative use of the latent distributions, as there is no unique solution to represent the data. 

Therefore, we use the generated embeddings to find a unique solution for the system by solving the Schr\"odinger equation with a perturbed potential \cite{Carleo602, han2018solving} . The embedded data $z_{i}$ is used to train a Neural Network in order to minimise the energies of our system according to the derivation explained in \ref{sec:Perturbation Theory}. 

The Neural Network used consists of a simple Multi-Layer Perceptron (MLP), with an $|Z| = d_{z}$ dimensional input and returning $\psi_{n}$ as output values. It is trained to optimise the following likelihood function 

\begin{gather*}
\label{eq:pert_energies_wavefunct}
 L(\psi , z) = E^{0} + E^{1} \\
 E^{0} = \alpha n^{2} \\
 \begin{split}
 E^{1} = (-1)^{t}\frac{cos(\pi t) [ 4n^{2} + t^{2} cos(2\pi n) - N^{2}] }{ }\\
 \frac{+ 2n [ t sin(2\pi n) sin (\pi t) - 2n]}{2\pi t [ t^{2} - 4n^{2}]}
 \end{split}
\end{gather*}

Where $\alpha$ weights the influence of the unperturbed or kinetic energy, and $t$ is an hyperparameter defined by the number of minima in the perturbation potential $V(z)$. The value of $n$, classically known as quantum number, can be derived from their respective periodic wave-functions, leading to $$n=\frac{L}{\pi} \frac{arc sin(\frac{L \psi^{2}}{2})}{z}$$


\section{Results}
\label{results}

The embeddings generated from the fully trained VAE provide a space where the clusters are clearly separable. This can be portrayed by using tSNE and other visualisation techniques over the latent dimensions, which depict the underlying topology of the generated space $Z$. But although these techniques are useful to visualise and interpret the results, their outputs are often non-generalisable, mostly due to the nature of their own algorithms. 
They don't account for the degeneration of the embeddings, making it very difficult to find the correlation between the latent space and the actual structure or distribution of the data observed. This is clearly detrimental for the goal of representation learning, where we ultimately would like to find the mapping between embeddings and the features or clustering structure of the data. 

Figure \ref{fig:energies_potential_schrodi} shows that the wave-functions or solutions of perturbation analysis derived in \ref{sec:Perturbation Theory} correspond with the spectrum of values generated by the MLP while optimising the perturbed Hamiltonian introduced in section \ref{implementation}. 

Each cluster therefore has a unique function $\psi_{n}$ with different quantum number $n$, and its corresponding energy $E_{n}$. The energy difference also indicates the cost of moving through the latent space and changing sample labels.

\section{Summary}
\label{summary}

We have developed a new approach to generalise the embeddings generated by VAEs. Perturbation theory from quantum mechanics can be used to solve the degeneration problem on the latent dimensions, providing a set of unique functions and energy spectrum that characterise the embedded space. 

We have shown that such energies and functions are related to the clusters or data structure, therefore can be generalised and used among several models or neural networks. 
The energy difference also accounts for topology and stability of the embedded data. The perturbation added to the Hamiltonian can also give an idea of the general energy landscape that outlines our system.  

This paper is only a first approach to the full potential of what combining theoretical physics and Machine Learning can deploy. Many problems currently faced by the deep learning community have already been tackled and solved by different branches of physics in the past. Therefore, we believe that by re-formulating and developing new approaches that combine both perspectives, one can better understand and build a robust framework for future progress in both fields.

\bibliography{paper_schrodi_VAE_final}
\bibliographystyle{icml2019}

%
%
%

\end{document}